\documentclass{article}

\usepackage{PRIMEarxiv}

\usepackage[utf8]{inputenc} 
\usepackage[T1]{fontenc}    
\usepackage{url}            
\usepackage{booktabs}       
\usepackage{amsfonts}       
\usepackage{nicefrac}       
\usepackage{microtype}      
\usepackage{lipsum}
\usepackage{fancyhdr}       
\usepackage{graphicx}       
\usepackage{amsmath}
\usepackage{amssymb}
\usepackage{caption}
\usepackage{multirow}
\usepackage{wrapfig}
\usepackage{subcaption}
\usepackage{url}            
\usepackage{xcolor}         
\usepackage[normalem]{ulem}
\usepackage{tabularx}
\usepackage{balance}

\usepackage[pagebackref,breaklinks,colorlinks]{hyperref}

\usepackage[capitalize]{cleveref}
\crefname{section}{Sec.}{Secs.}
\Crefname{section}{Section}{Sections}
\Crefname{table}{Table}{Tables}
\crefname{table}{Tab.}{Tabs.}

\title{Visualizing Transferred Knowledge: An Interpretive Model of Unsupervised Domain Adaptation
}

\author{
  Wenxiao Xiao \\ 
  Department of Computer Science\\
  Brandeis University\\
  \texttt{wenxiaoxiao@brandeis.edu} \\
  \And
  Zhengming Ding \\
  Department of Computer Science \\
  Tulane University \\
  \texttt{zding1@tulane.edu} \\
  \And
  Hongfu Liu \\
  Department of Computer Science\\
  Brandeis University\\
  \texttt{hongfuliu@brandeis.edu} \\
}

\begin{document}
\maketitle

\begin{abstract}
Many research efforts have been committed to unsupervised domain adaptation (DA) problems that transfer knowledge learned from a labeled source domain to an unlabeled target domain. Various DA methods have achieved remarkable results recently in terms of predicting ability, which implies the effectiveness of the aforementioned knowledge transferring. However, state-of-the-art methods rarely probe deeper into the transferred mechanism, leaving the true essence of such knowledge obscure. Recognizing its importance in the adaptation process, we propose an interpretive model of unsupervised domain adaptation, as the first attempt to visually unveil the mystery of transferred knowledge. 
Adapting the existing concept of the prototype from visual image interpretation to the DA task, our model similarly extracts shared information from the domain-invariant representations as prototype vectors. Furthermore, we extend the current prototype method with our novel prediction calibration and knowledge fidelity preservation modules, to orientate the learned prototypes to the actual transferred knowledge.
By visualizing these prototypes, our method not only provides an intuitive explanation for the base model's predictions but also unveils transfer knowledge by matching  the image patches with the same semantics across both source and target domains. Comprehensive experiments and in-depth explorations demonstrate the efficacy of our method in understanding the transferred mechanism and its potential in downstream tasks including model diagnosis. 
\end{abstract}


\section{Introduction}\label{sec:intro}

Unsupervised domain adaptation (DA) aims to transfer knowledge learned from a well-labeled domain to another target domain with no annotation. By projecting different domains to a domain-invariant representation, recent DA methods~\cite{Chen_2019_CVPR, dosovitskiy2020image} bring off remarkable results in predicting the unlabeled data. The success of DA methods~\cite{wang2020unsupervised, yang2021tvt, xu2021cdtrans} evinces that valid source information gets transferred to the target domain during the adaptation, but the nature of this transferred knowledge is yet to be explored. 

This lack of understanding of the knowledge transfer in DA models might raise doubts on their real-life applications especially when the stakes are high. In addition, being able to decipher the transferred knowledge also helps in downstream tasks including model diagnosis and improvement. Regrettably, current DA methods\cite{9879990, ganin2016domain, liang2020baus} mostly approach such knowledge implicitly and can only be explained to a limited extent, i.e., sample-level weights~\cite{jin2020minimum}. Although considerable interpretive methods~\cite{li2021interpretable,zhang2021survey} have been proposed to identify the most important input features~\cite{hatt2019machine,simonyan2013deep} that lead to the predictions, few can provide sufficient insight into the transferring process. Specifically, nearly none of them helps intuitively understand the transfer mechanism and its effect on existing DA models. The absence of efforts to interpret the transferred information motivates us to fill in the blanks, considering its principal role in DA. 

In this paper, we propose our visual interpretive model for unsupervised domain adaptation to shine a light on transferred knowledge. Given a well-trained DA model, our model extracts information from the domain-invariant feature space as category-specific prototypes~\cite{NEURIPS2019_adf7ee2d} with a knowledge extraction module. These prototypes are associated with source latent patches of their assigned category, and thus can be visualized by image parts of source samples. As both domains are projected onto the same representations, the learned prototypes can extract various shared knowledge. 
To ensure the prototypes attend to this shared information, we train our interpretative model to predict based on the samples' similarity to the learned prototypes with a prediction calibration module. Then our knowledge fidelity preservation module further aligns the predicted distribution of our model with the base DA model, leading the prototypes to grasp the transferred knowledge. 
Each prototype matches similar image parts across domains with its semantic information, providing a visual association between the source and target samples, where the knowledge transfer can be explained as \textit{``This target image part looks like that source sample part since they share the same semantics."}

\noindent \textbf{Contribution.} Our major contributions are summarized in three folds as follows:
\begin{itemize}
\vspace{-2mm}
    \item We propose an interpretative method for unsupervised domain adaptation to visualize the transferred knowledge. To the best of our knowledge, this is the first attempt to explain DA models with the cross-domain connection established on the transferred knowledge.
\vspace{-2mm}    
    \item We adapt the existing prototype framework to solve this crucial yet under-researched problem in DA tasks. Specifically, we extend the ProtoPNet~\cite{NEURIPS2019_adf7ee2d} with our novel knowledge calibration modules, providing a case-based visual interpretation of transferred knowledge.
\vspace{-2mm}    
    \item We exhibit the efficacy of our model with comprehensive experiments on two benchmark datasets and validate our interpretation of the base DA model with  various in-depth analyses and explicit examples. 
\end{itemize}

\section{Related works}\label{sec:related}
\noindent \textbf{Interpretative models.}
Interpretative models\cite{li2021interpretable,zhang2021survey} aim to provide certain explanations for the black-box deep networks. The vast majority of these methods fall into two genres, rule-based~\cite{pmlr-v97-goyal19a,10.5555/1625855.1625918} and feature attribution methods~\cite{simonyan2013deep, baehrens2010explain}. Researchers working in the first category attempted to extract rule-based interpretations on the global level in their early works. 
For example, works including \cite{quinlan2014c4, odajima2008greedy} interpret a deep model with a set of rules~\cite{nayak2009generating} or a decision tree~\cite{krishnan1999extracting} extracted from model-generated samples. Other works focus on explaining the prediction of a single sample with local-level logic rules. CDRPs~\cite{wang2018interpret} locally explains a model's behavior by isolating the critical nodes and connections in the network when the model makes individual predictions. CEM~\cite{dhurandhar2018explanations}, CVE~\cite{pmlr-v97-goyal19a} and DACE~\cite{kanamori2020dace} identify the critical parts of an individual sample by finding permutation of the input features that produce the same output. 
Alternatively, a significant amount of attribution methods explain the deep model by assigning importance scores to the input features after Explanation Vectors\cite{baehrens2010explain} attempts to open the black-box models with gradients. Following this direction, several works~\cite{zeiler2014visualizing, springenberg2014striving} calculate gradients-based salient maps for understanding computer vision models. Apart from gradient-based attribution, other efforts including LIME~\cite{ribeiro2016should} and SHAP~\cite{lundberg2017unified} try to obtain attribution in a model-agnostic way. Recently, some methods~\cite{guo2019deep, hatt2019machine} combine local attributions for a global model-agnostic interpretation.

\noindent \textbf{Visual image interpretation.}
In general, research efforts devoted to visual image interpretation can be divided into two groups, \textit{post-hoc} and \textit{self-interpretable}. Post-hoc methods~\cite{pmlr-v97-goyal19a,simonyan2013deep} open the black-box models with salient maps or key image parts after the deep network is finished training. Most of the interpretative models~\cite{zeiler2014visualizing, springenberg2014striving} we discussed in the previous paragraph fall into this group, locally explaining a model's prediction for each individual sample. Extending these gradient-based methods, some works~\cite{bach2015pixel, shrikumar2017learning, selvaraju2017grad} include various additional information to achieve more reasonable explanations. On the global level, TCAV~\cite{kim2018interpretability} constructs a set of explanatory concepts represented by vectors that are able to separate positive/negative examples in a hidden layer. On the other hand, self-interpretable\cite{plumb2020regularizing, weinberger2020learning} methods generate explainable representations in an end-to-end training process. Recently, ProtoPNet~\cite{NEURIPS2019_adf7ee2d} proposed a self-interpretable global interpretation method that extracts category-specific prototype vectors associated with image patches of training samples, which serve as an example-based explanation on both global and local levels. Later, various methods~\cite{nauta2021neural, hase2019interpretable} extend ProtoPNet from different aspects, including refined prototype learning module~\cite{donnelly2022deformable, Wang_2021_ICCV} and pruning strategies~\cite{rymarczyk2021protopshare} to reduce the number of prototypes.

\noindent \textbf{Unsupervised domain adaptation.}
To align the heterogeneous distributions in source and target domains, previous works focusing on domain adaptation can be roughly separated into two straits. Early efforts~\cite{luo2014decomposition, pmlr-v48-gong16, pan2010domain} alleviate the discrepancy by matching certain high-order statistics of a different domain, which essentially aligns the distribution on the domain level. For example, maximum mean discrepancy (MMD)~\cite{borgwardt2006integrating} is one of the widely-used statistics in the domain-level adaptation methods~\cite{tzeng2014deep, long2015learning}. Recently, methods including AFN~\cite{xu2019larger} proposed various new statistics to improve the adaptation. The other strait of the DA methods~\cite{pei2018multi, long2018conditional}, inspired by Generative Adversarial Networks~\cite{NIPS2014_5ca3e9b1}, resorts to learning a domain-invariant representation that can trick an adversarial domain discriminator. Extending the initial DANN~\cite{ganin2016domain}, following methods~\cite{Chen_2019_CVPR, wang2020unsupervised} adopt some re-weights schema to align different distributions on the category level. Lately, NWD~\cite{9879990} introduces Nuclear-norm Wasserstein discrepancy, which allows the original task-specific classifier to be reused as a discriminator for further aligning two domains. After the vision transformer~\cite{dosovitskiy2020image} is introduced, several works~\cite{yang2021tvt, xu2021cdtrans} incorporate this new architecture and achieve promising results. However, although some state-of-arts provide certain interpretations of their model, mostly with the weights on the categories or samples, none of them dig into the adaptation process to unveil the transferred knowledge. 

\begin{figure*}\vspace{-8mm}
  \centering
  \includegraphics[width=\textwidth]{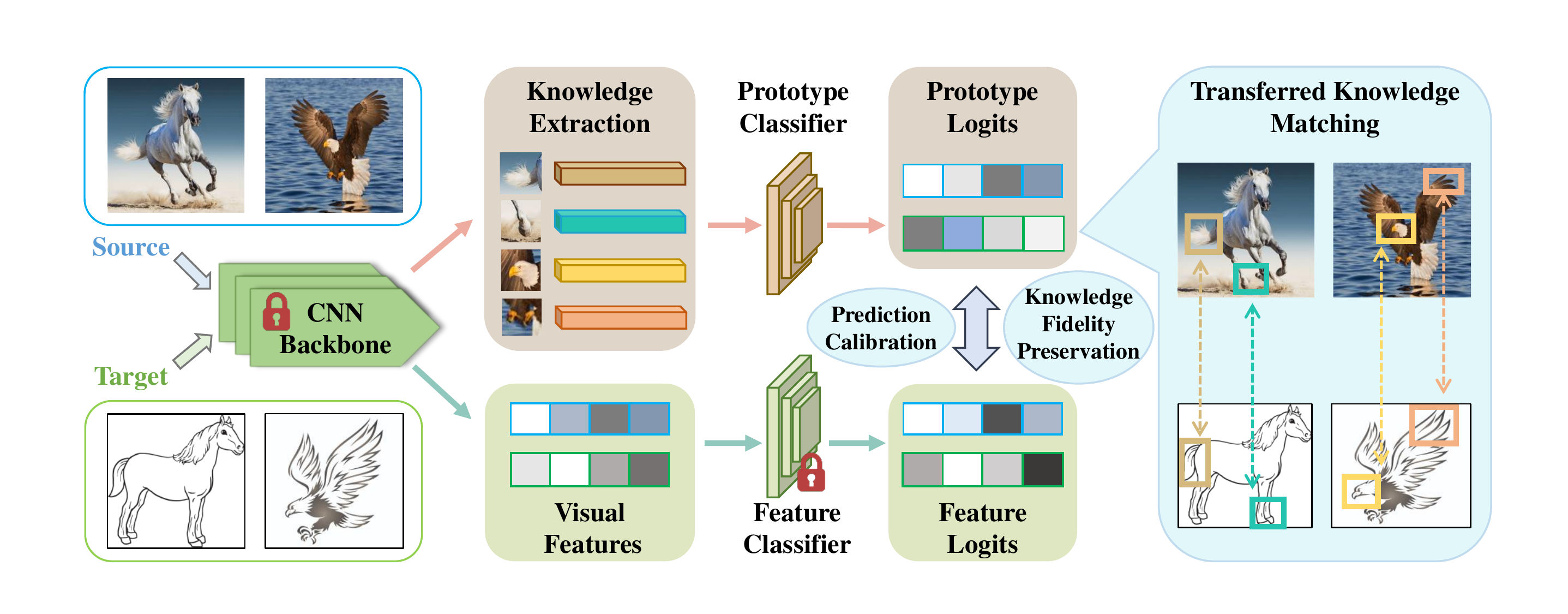}\vspace{-4mm}
  \caption{Framework of our proposed transferred knowledge visualization in unsupervised domain adaptation. The knowledge extraction module extricates category-specific prototypes that represent various transferred knowledge from the base model's domain-invariant feature space. The prediction calibration module emulates the fixed feature classifier's decision of a sample image based on its similarities to the extracted prototypes. The knowledge fidelity preservation further aligns the soft-max distributions of the outputs of two classifiers, encouraging the prototypes to learn from the transferred knowledge utilized by the base model. The rightmost part demonstrates how the learned prototypes visually match image parts corresponding to the transferred knowledge across the source and target domains.}\vspace{-4mm}
  \label{fig:framework}
\end{figure*}

\section{Problem formulation}\label{sec:formulation}
With all endeavors committed to opening the black-box neural networks, researchers in this field rarely set their sights on explaining the transferring process of unsupervised domain adaptation. In fact, current state-of-the-art CNN-based DA methods~\cite{9879990, wang2020unsupervised, li2021semantic} achieve impressive results on the benchmark datasets, indicating these methods indeed transfer their knowledge learned from the domain-invariant features. However, there is nearly no dedicated effort to reveal the \textit{transferred knowledge}, despite the fact that the community has been discussing transferring between different domains for quite a long time. 
In fact, interpreting the transferred information utilized by the base DA methods is of great value, especially if we can connect the source and target domains with such knowledge. Being able to recognize the meaningful transferred knowledge allows users to rely on the model with confidence while isolating the incorrectly transferred information facilities model diagnosis and potential downstream tasks like semantic learning. In light of this, a natural question arises with the significant advance in unsupervised domain adaptation:
\vspace{-2mm}
\begin{center}
\textit{\textbf{What does transferred knowledge look like in an unsupervised domain adaptation model, and \\how does such knowledge facilitate the adaptation?}}
\end{center}
\vspace{-2mm}
Unfortunately, the state-of-the-art DA models apprehend the transferred knowledge in an implicit way, i.e., re-weighting the training samples~\cite{jin2020minimum}, which does not help understand its underlying meaning. Meanwhile, with all the distinguished interpretative methods, they focus on interpreting their predictive results but nearly none of them provides sufficient explanations for the transferred knowledge in unsupervised domain adaption. Therefore, we would like to take the initiative by proposing our interpretative method of unsupervised DA for visualizing transferred knowledge.

\section{Methodology}
In this section, we elaborate on our proposed transferred knowledge visualization model. We first give an overview of the framework. Then we discuss the components and their corresponding objective functions of our model in detail.

\subsection{Framework overview}
The framework of our visualizing transferred knowledge in unsupervised domain adaptation is demonstrated in \Cref{fig:framework}, which consists of three components, namely, (i) knowledge extraction, (ii) prediction calibration, and (iii) knowledge fidelity preservation. The knowledge extraction aims to uncover knowledge from the domain-invariant feature representation of the base DA model, while the prediction calibration encourages our interpretative model to make similar decisions as the original one based on the uncovered knowledge. In addition to imitating the final predictions, the knowledge fidelity preservation further ensures that the extracted knowledge indeed represents the information utilized during the transfer process, by aligning the soft-max distributions coming out of two classifiers. The assorted semantics within the transferred knowledge is then visualized with image patches across the source and target domains. By matching image patches corresponding to the same semantics across domains, our model interprets transferred knowledge in a straightforward way: \textit{The knowledge of this semantics is transferred from those source samples to this target image, as the same prototype matches them together.}

Inspired by ProtoPNet~\cite{NEURIPS2019_adf7ee2d}, which explains objective recognition models with prototypical image examples, our knowledge extraction module learns prototype vectors associated with image parts of the source samples from the pre-trained features. Each prototype is assigned to one category to catch the category-specific knowledge shared by both domains. The similarity scores between each sample and these prototypes then serve as the basis for making predictions. The prediction calibration module strives to mirror the base model's behavior by training the classifier of our interpretative model against the predictions of the base DA model instead of the ground truth, as our motivation is to explain the DA model's decision, even when it defies the ground truth. Still, we are one step away from our final goal, since the same prediction cannot guarantee the prototypes contain the transferred knowledge. Our method narrows down the gap with knowledge fidelity preservation, which aligns the output distributions of the frozen feature classifier and our prototype classifier, forcing the prototypes to learn the information transferred by the DA model. These prototypes constitute a case-based global interpretation for the base model. Moreover, our model provides visual cues of the transferred knowledge by matching each prototype to the image parts of its corresponding category across both domains, which can help users to interpret the transferring process and diagnose the base DA model.

\subsection{Preliminaries} 

Given a source domain $\mathcal{D}^s$$=$$\{(x^s_1, y^s_1),..., (x^s_{n_s}, y^s_{n_s})\}$ with $n_s$ samples from $\mathcal{C}$$=$$\{C_1, ... , C_c\}$ categories and an unlabelled target domain $\mathcal{D}^t$$=$$\{x^t_1,...,x^t_{n_t} \}$ from $\mathcal{C}$ categories as the source domain. In general, a CNN-based unsupervised domain adaptation method projects each sample image $x$ onto a 3D-feature space with the dimension of $H$$\times$$W$$\times$$D$, with a backbone feature extractor $f$, e.g., ResNet~\cite{he2016deep} or VGG~\cite{simonyan2014very}. Then the feature spaces of both domains are aligned together and merged into one domain-invariant representation space with various re-weighting schema. This shared representation space enables the feature classifier $h_f$, which is trained on the source domain supervised by ground truth, to also make predictions for the target samples, implicitly transferring knowledge between domains. 
Similar to ProtoPNet~\cite{NEURIPS2019_adf7ee2d}, we employ a prototype layer $g$ to identify the informative image parts for each category with $K$ prototype vectors $P^k$$=$$\{p_i^k|_{k=1}^K\}$ of size $H'$$\times$$W'$$\times$$D$ for the $i$-th individual category $C_i$, where $H'$$<$$H$ and $W'$$<$$W$. The maximum $L^2$ similarity scores of $x$ to each of the $K$$\times$$c$ prototype vectors, donated as $g \circ f(x)$, is calculated as the input of our interpretative prototypical classifier $h_p$ for prediction. 

\begin{table*}[!t]
  \caption{Our interpretation on DANN for Close-set Domain Adaptation on \textit{Office-Home} and \textit{DomainNet-126} by predicting accuracy (\%)}\vspace{-2mm}
  \label{table:dann_OfficeHome}
  \linespread{1.3} 
  \centering 
  \resizebox{1.\textwidth}{!}{
  \begin{tabular}{lcccccccccccccc}
    \toprule
    \textit{Office-Home} &Ar→Cl &Ar→Pr &Ar→Rw &Cl→Ar &Cl→Pr &Cl→Rw &Pr→Ar &Pr→Cl &Pr→Rw &Rw→Ar &Rw→Cl &Rw→Pr &Avg. \\
    \midrule
    DANN~\cite{ganin2016domain}  &53.26 &60.57 &70.53 &53.93 &63.70 &65.61 &54.92 &54.73 &75.87 &66.33 &60.64 &78.40 &63.20 \\
    \midrule
    Ours                         &53.22 &60.55 &70.74 &53.44 &63.71 &65.07 &54.63 &54.02 &75.60 &66.25 &59.98 &78.58 &62.99\\
    Diff. from DANN              &-0.04 &-0.02 &+0.21 &-0.49 &+0.01 &-0.54 &-0.29 &-0.71 &-0.27 &-0.08 &-0.56 &+0.18 &-0.21\\
    \midrule
    \midrule
    \textit{DomainNet-126}                          &Cl→Pt   &Cl→Rl   &Cl→Sk   &Pt→Cl  &Pt→Rl   &Pt→Sk   &Rl→Cl   &Rl→Pt   &Rl→Sk   &Sk→Cl   &Sk→Pt   &Sk→Rl   &Avg. \\
    \midrule
    DANN~\cite{ganin2016domain}                        &47.86 &60.57 &52.24 &55.10 &68.22 &53.64 &61.62 &62.53 &55.21 &61.30 &57.34 &63.29 &58.24 \\
    \midrule
    Ours                        &42.58 &56.93 &47.38 &53.48 &65.77 &49.38 &57.60 &57.76 &51.02 &56.14 &51.11 &58.75 &54.00\\
    Diff. from DANN             &-5.28 &-3.64 &-4.86 &-1.62 &-2.45 &-4.26 &-4.02 &-4.77 &-4.19 &-5.16 &-6.23 &-4.54 &-4.24\\
    \bottomrule
  \end{tabular}
  }\vspace{-4mm}
\end{table*}

\subsection{Transferred knowledge recognition}
\noindent \textbf{Knowledge extraction.} We adopt ProtoPNet's prototype layer to extract prototypes from the shared feature space $f(x)$. The knowledge extraction includes two loss terms, i.e., the clustering loss $\mathcal{L}_c$ and the separation loss $\mathcal{L}_s$. $\mathcal{L}_c$ encourages each prototype $p_i^k$ to be similar to some latent patch for source samples of its assigned category $C_k$, while $\mathcal{L}_s$ tries to push $p_i^k$ away from all the latent patches coming out of $C_k$ in the source domain. This component allows our model to extract category-specific transferable prototypes, as the domain-invariant feature space is pre-aligned by the base model. Technically, we define these two loss terms:
\begin{equation} \label{eq:L_c_s}
\mathcal{L}_{c}  = \frac{1}{n_s}\sum_{i=1}^{n_s}\underset{j:p_j \in P^{y_i}}{\operatorname{\min}}\underset{z \in f(x_i)}{\operatorname{\min}}||z - p_j||^2\textrm{, and } \mathcal{L}_{s} = -\frac{1}{n_s}\sum_{i=1}^{n_s}\underset{j:p_j \notin P^{y_i}}{\operatorname{\min}}\underset{z \in f(x_i)}{\operatorname{\min}}||z - p_j||^2,
\end{equation} where $z$ is the latent patch of $f(x_i)$ with the same dimension to prototype vector.

\noindent \textbf{Prediction calibration.} As an interpretative method, our model does not seek to achieve a higher performance in terms of predicting accuracy compared with the ground truth. Instead, we want our model makes the same decisions for both source and target domains as the base DA model based on the transferable prototypes. Thus, we do not involve the ground truth labels during the training, but supervise the training of our interpretative classifier $h_p$ with the base DA model's predicted labels. Mathematically, this module can be expressed by the following cross-entropy loss:
\begin{equation} \label{eq:L_cls}
\mathcal{L}_{Cls} =  \frac{1}{n_s}\sum_{i=1}^{n_s}\ell_{\textup{CE}}(h_p \circ g \circ f(x^s_i), \hat{y}^s_i)
                  +  \frac{1}{n_t}\sum_{i=1}^{n_t}\ell_{\textup{CE}}(h_p \circ g \circ f(x^t_i), \hat{y}^t_i),
\end{equation} where $\ell_{\textup{CE}}$ denotes the cross-entropy loss, and $\hat{y}^s_i$ and $\hat{y}^t_i$ are the predicted labels of the base classifier $h_f$, as we want to calibrate the interpretative classifier $h_p$ to $h_f$.


\noindent \textbf{Knowledge fidelity preservation.} The above components encourage our method to mimic the base DA model with prototypes shared by both domains. However, to achieve our goal, which is interpreting the transferred knowledge, getting the same predicted labels is not enough. To bridge the gap, we employ knowledge fidelity preservation to capture the information that is actually get transferred by the base model. Technically, this preservation step encourages the prototypical classifier $h_p$ to imitate the original feature classifier $h_f$ as much as possible, by enforcing an $L_1$ loss on the softmax logits between the output of $h_p$ and $h_f$ on the target domain, which can be expressed as:
\begin{equation} \label{eq:L_fid}
\mathcal{L}_{Fid} =\frac{1}{n_t}\sum_{i=1}^{n_t} ||h_p \circ g \circ f(x_i^t) - h_f \circ f(x_i^t) ||_1,
\end{equation}
where $h_p \circ g \circ f(x_i^t) \in \mathbb{R}^c$ and $h_f \circ f(x_i^t) \in \mathbb{R}^c$ are the \textit{softmax} predictions of the fixed base classifier and the prototypical classifier for target sample $x_i$, respectively. 

Different from $\mathcal{L}_{Cls}$ that specifically punishes the incorrect prediction with a cross-entropy loss, this fidelity loss intends to align the softmax distributions from two classifiers, pushing the prototypes to exploit the transferred knowledge.

\subsection{Training protocol and overall objectives} Our model is a post-hoc method as its training process requires a pre-trained base DA model. With the pre-trained feature extractor $f$ and classifier $h_f$, we train our interpretive model by rotating three stages.

\noindent \textbf{Optimizing prototype layer.} In this stage, we want to extract prototypes from the domain-invariant feature space that is not only closely associated with their designated categories and well-separated from other categories but also contains the transferred knowledge during adaptation. Thus, we refine the prototypical classifier $h_p$ and train the prototype layer $g$ with the following objective function:
\begin{equation} \label{eq:beforelast}
\underset{g}{\operatorname{\min}}\ \mathcal{L}_{Cls} + \alpha\mathcal{L}_{c} + \beta\mathcal{L}_{s} + \gamma\mathcal{L}_{Fid},
\end{equation}
where $\alpha$, $\beta$, and $\gamma$ are the trade-off parameters. To avoid a cold start, we do not randomly initialize the weights in $h_p$. Instead, we set the weights between each prototype to its assigned category to 1, and the other weights are all set to -0.5 at the beginning of the training.  

\noindent \textbf{Prototype projection.} In order to visualize knowledge carried in the prototype vectors in the form of sample image patches, we project each prototype vector $p_i^k$ back to the nearest latent patch across all the source samples of its corresponding category. This projection step further ensures that all the prototypes represent certain information from the source images. Moreover, as the prototypes are extracted from the domain-invariant feature space, we believe the projected prototypes carry the transferred knowledge that bridges the two domains. This projection is defined as:
\begin{equation} \label{eq:projection}
p_j^k \Leftarrow \space \underset{z \in f(X^s_k)}{\operatorname{ arg\min }}||z-p_j^k||^2,
\end{equation}
where $f(X^s_k)$ are all latent patches of source category $C_k$.

\noindent \textbf{Optimizing the last layer.} In this stage, we calibrate our interpretive classifier $h_p$ with the classifier of base DA by fixing the prototype layer $g$. With the help of fidelity preservation, $h_p$ is trained to mimic the base classifier $h_f$ to the greatest extent. As a result, $h_p$ will encourage the prototypes to uncover the transferred knowledge used by the base DA model in the next round of prototype layer training. The stage can be expressed as:
\begin{equation} \label{eq:last}
\underset{h_p}{\operatorname{\min}}\ \mathcal{L}_{cls} + \lambda||W_{h_p}||_1,
\end{equation}
where $W_{h_p}$ is the collection of weights in the classifier $h_p$,  and $\lambda$ is the trade-off parameter for the $L_1$ regularizing term.

\subsection{Visualization matching}
Similar to ProtoPNet~\cite{NEURIPS2019_adf7ee2d}, we visualize the learned prototypes with training image parts based on the similarity. Globally, to visualize a prototype $p$, we find the projected source image $x$ for each prototype as in \cref{eq:projection} and calculate the $L_2$ similarity heatmap between $f(x)$ and $p$. Based on the upsampled similarity heatmap, we visualize $p$ with the most activated parts, which are cropped out with the smallest bounding box that contains all locations with the highest 95\%. As we discussed in \Cref{sec:formulation}, our motivation is to visualize the transferred knowledge and match the image parts across two domains with this knowledge. Thus, for the source and target images $x^s_i$ and {$x^t_i \in C_j$}, we use the same strategy to crop out the image parts in both $x^s_i$ and {$x^t_i$} for each of the $K$ prototypes assigned to class $C_j$ and visualize how the transferred knowledge connect two domains locally on the sample level. With this visualization matching, we can examine what information is successfully transferred to the target domain, and identify the misaligned prototypes that impair the adaptation for diagnosis purposes.

\begin{figure*}
  \centering
  \includegraphics[width=\textwidth]{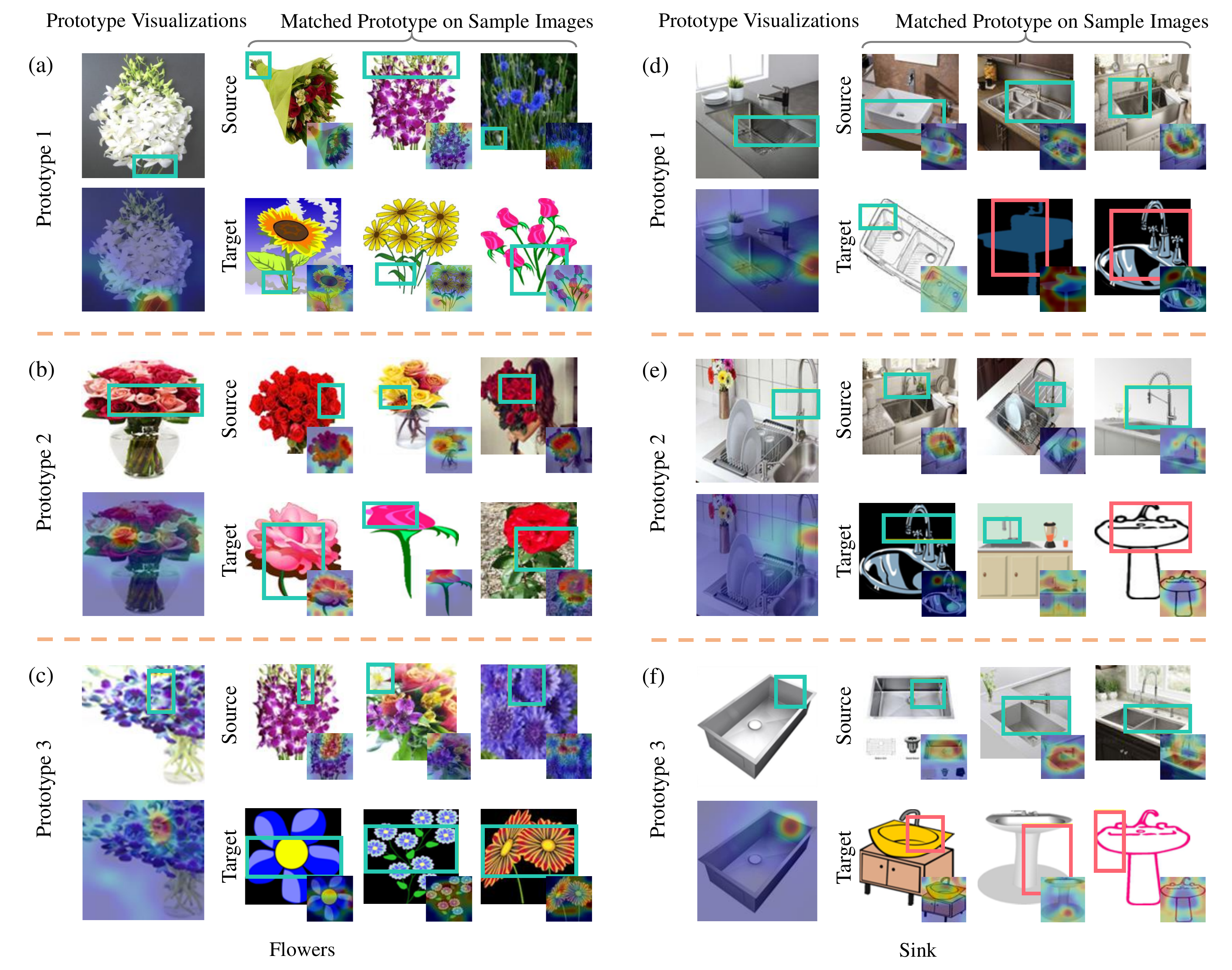}\vspace{-2mm}
  \caption{Examples for prototype visualization and transferred knowledge matching of {Flowers} (left) and {Sink} (right) categories in task Pr→Cl on \textit{Office-Home} dataset. The image patches corresponding to each detected prototype are cropped out with rectangular bounding boxes for both source and target samples. The mismatched target image patches are highlighted with red bounding boxes.}
  \label{fig:matching}\vspace{-3.5mm}
\end{figure*}

\section{Experiments}
\subsection{Experimental setup}
\noindent \textbf{Datasets}. We choose two popular DA benchmark datasets, \textit{Office-Home}~\cite{venkateswara2017deep} and  \textit{DomainNet-126}~\cite{peng2019moment} in our experiments. \textit{(i) Office-Home}~\cite{venkateswara2017deep} is one of the most popular DA benchmark datasets, which contains images of 65 different categories from four domains: Art (Ar), Clip Art (Cl), Product (Pr), and Real World (Rw). We include all 12 available adaptation tasks in our experiments. \textit{(ii) DomainNet-126} is a subset of \textit{DomainNet}~\cite{peng2019moment}, the current largest DA benchmark dataset, which consists of 345 categories from 6 different domains. We follow the same data protocol as \cite{saito2019semi} and choose 126 categories from 4 different domains: Real (Rl), Clip Art (Cl), Painting (Pt), and Sketch (Sk) since the labels for certain domains and categories are very noisy. We also conduct all possible adaptations for these 4 selected domains.

\begin{figure*}[t]
     \begin{subfigure}[b]{0.31\textwidth}
         \centering
         \includegraphics[width=\textwidth]{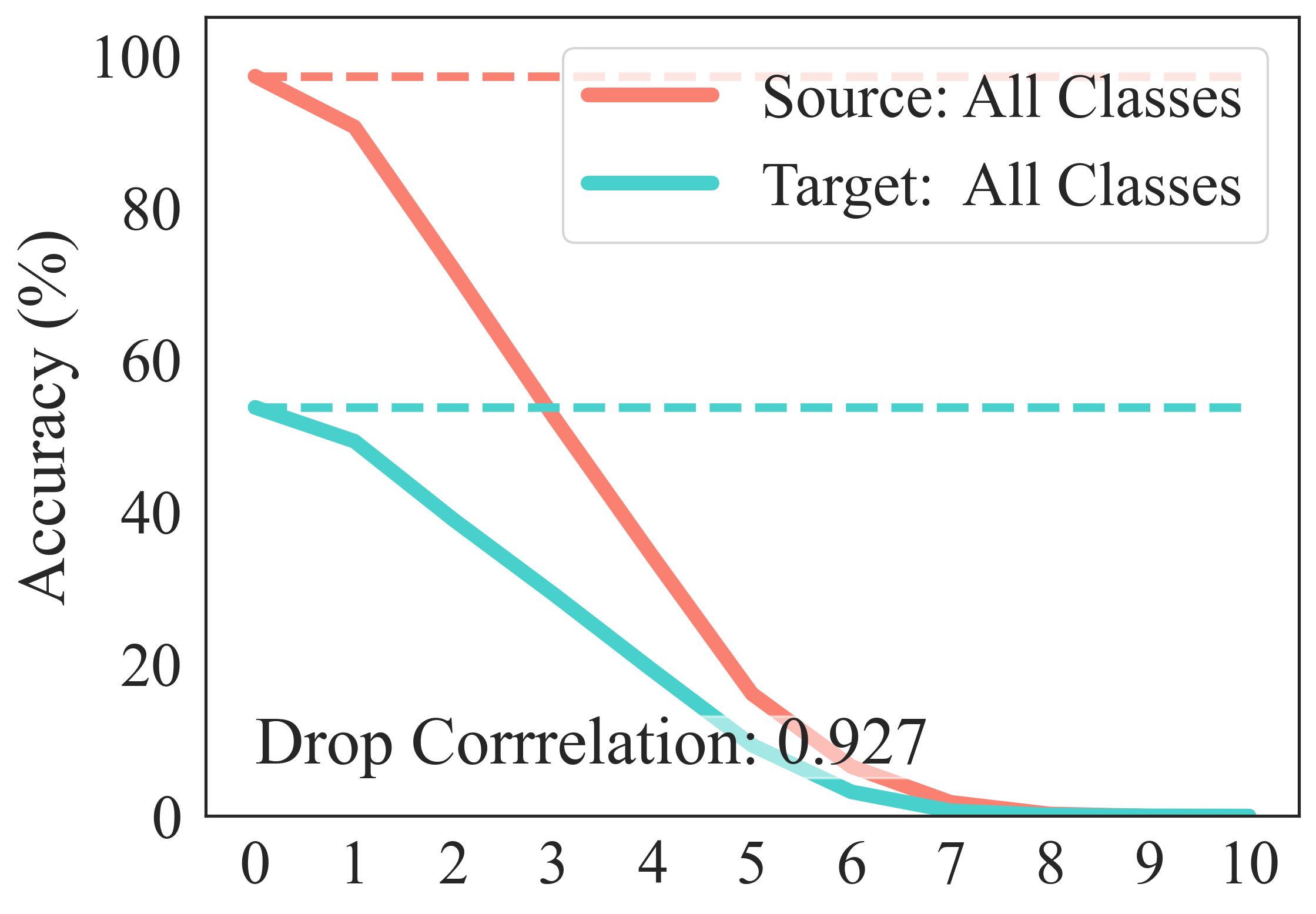}
         \caption{ \hspace{0.1cm} Pr→Cl:\hspace{0.1cm}  All Classes}
         \label{fig:Pr2Cl}
     \end{subfigure}
     \hfill
     \begin{subfigure}[b]{0.31\textwidth}
         \centering
         \includegraphics[width=\textwidth]{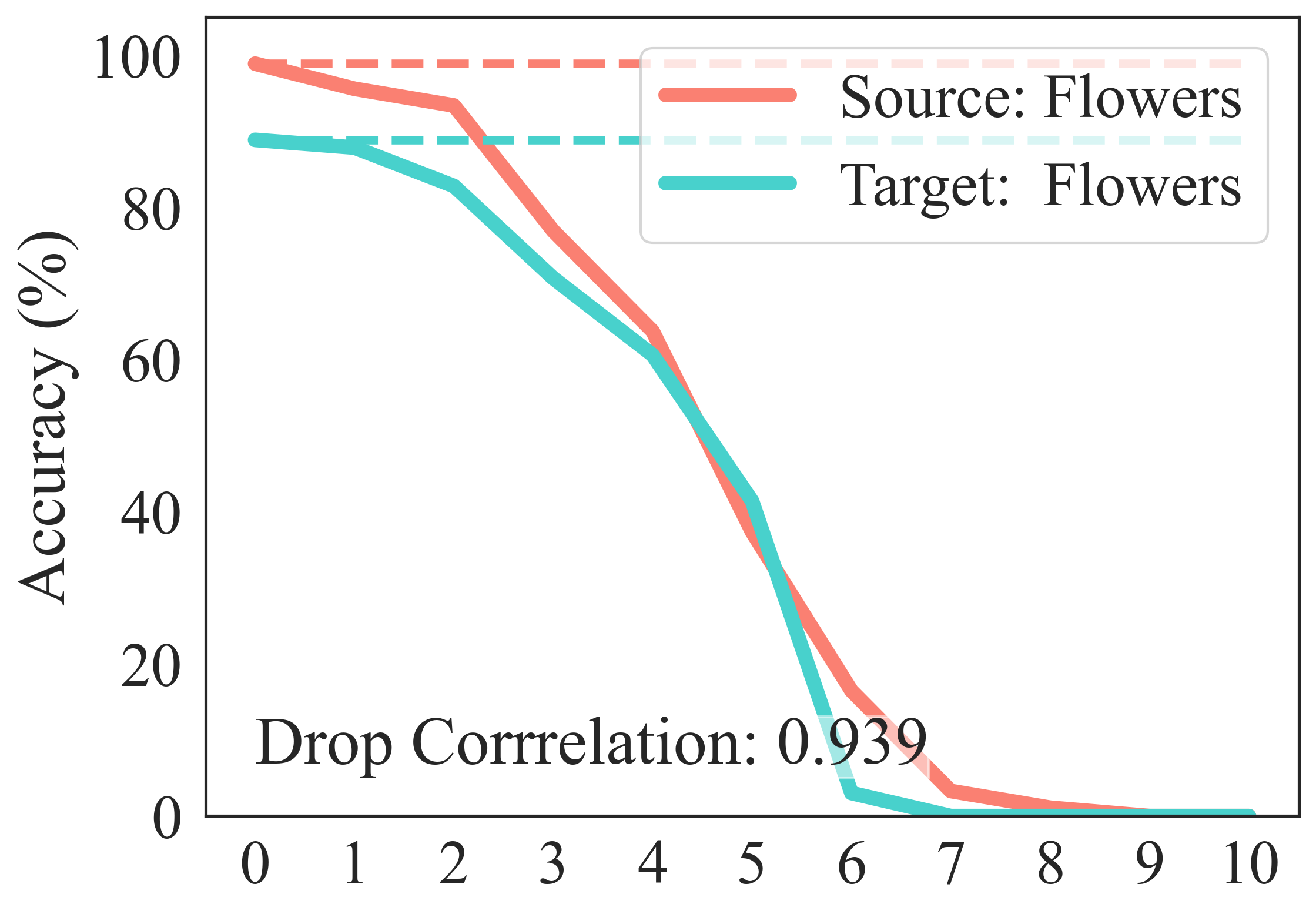}
         \caption{ \hspace{0.1cm} Pr→Cl: \hspace{0.1cm} Flowers}
         \label{fig:Pr2ClFlowers}
     \end{subfigure}
     \hfill
     \centering
     \begin{subfigure}[b]{0.31\textwidth}
         \centering
         \includegraphics[width=\textwidth]{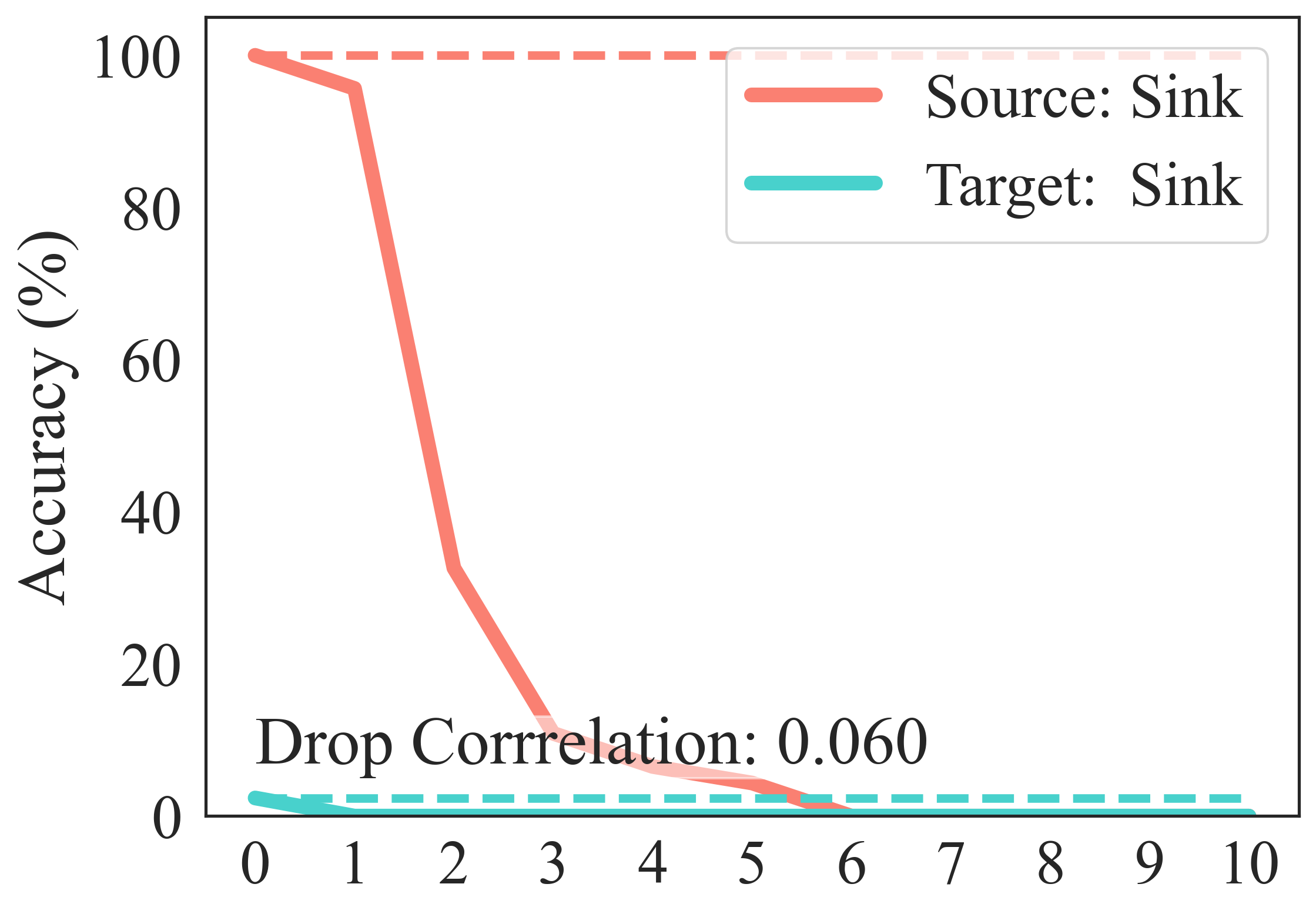}
         \caption{ \hspace{0.1cm} Pr→Cl: \hspace{0.1cm} Sink}
         \label{fig:Pr2ClSink}
     \end{subfigure}
     \hfill
     \begin{subfigure}[b]{0.31\textwidth}
         \centering
         \includegraphics[width=\textwidth]{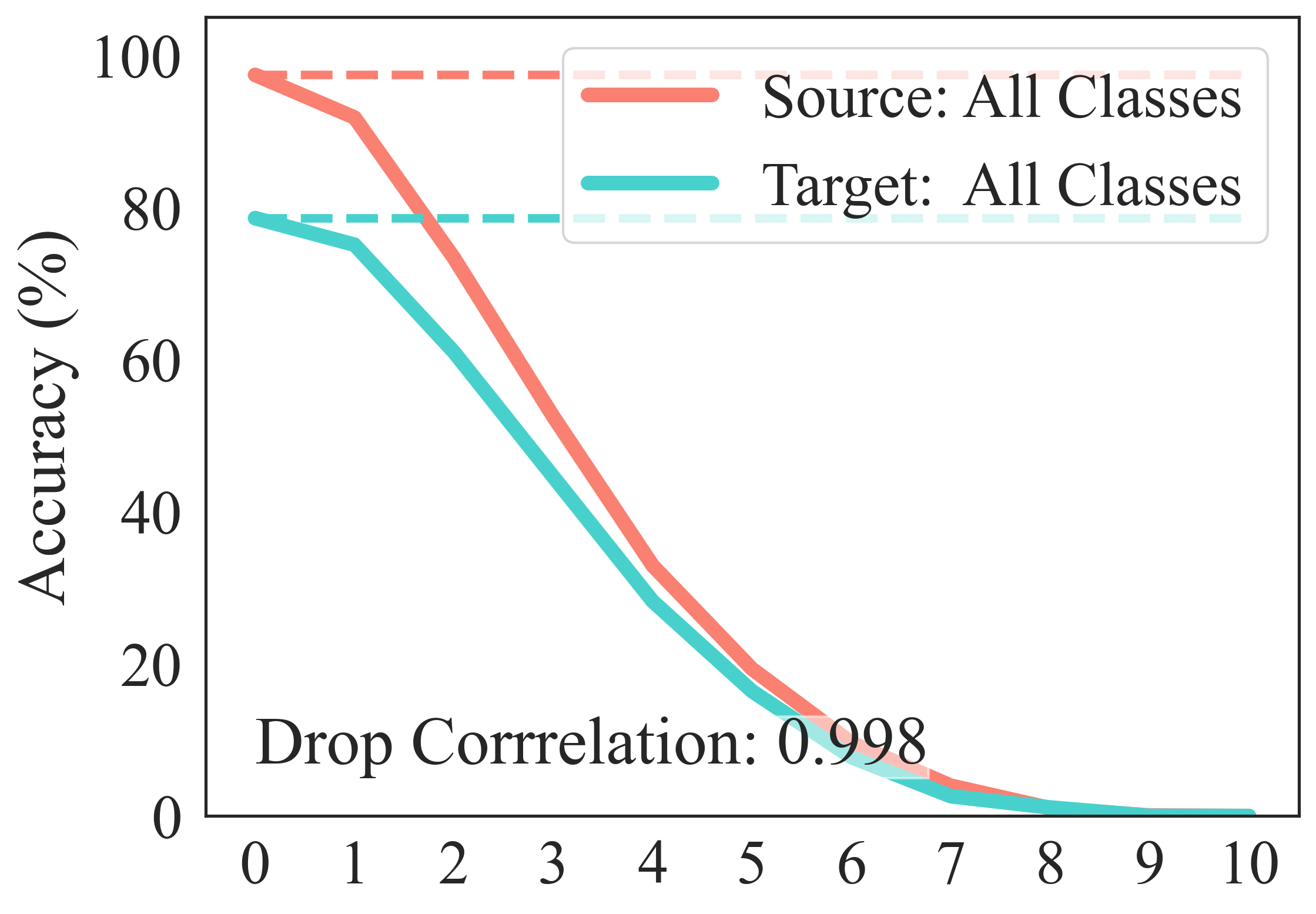}
         \caption{ \hspace{0.1cm} Rw→Pr: \hspace{0.1cm}  All Classes}
         \label{fig:Rw2Pr}\vspace{-1mm}
     \end{subfigure}
     \hfill
     \begin{subfigure}[b]{0.31\textwidth}
         \centering
         \includegraphics[width=\textwidth]{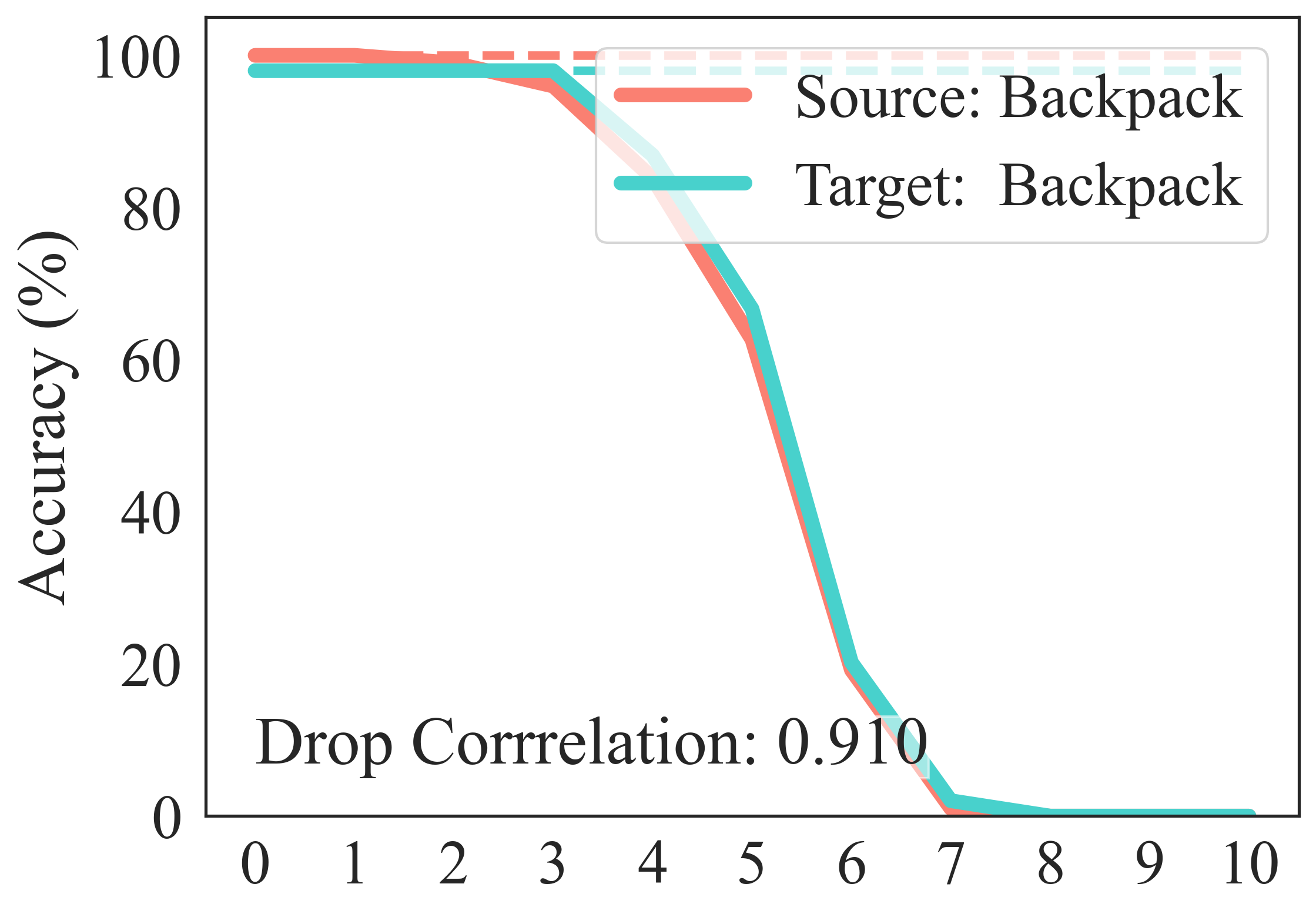}
         \caption{ \hspace{0.1cm} Rw→Pr: \hspace{0.1cm} Backpack}
         \label{fig:Rw2PrBackpack}\vspace{-1mm}
     \end{subfigure}
     \hfill 
     \begin{subfigure}[b]{0.31\textwidth}
         \centering
         \includegraphics[width=\textwidth]{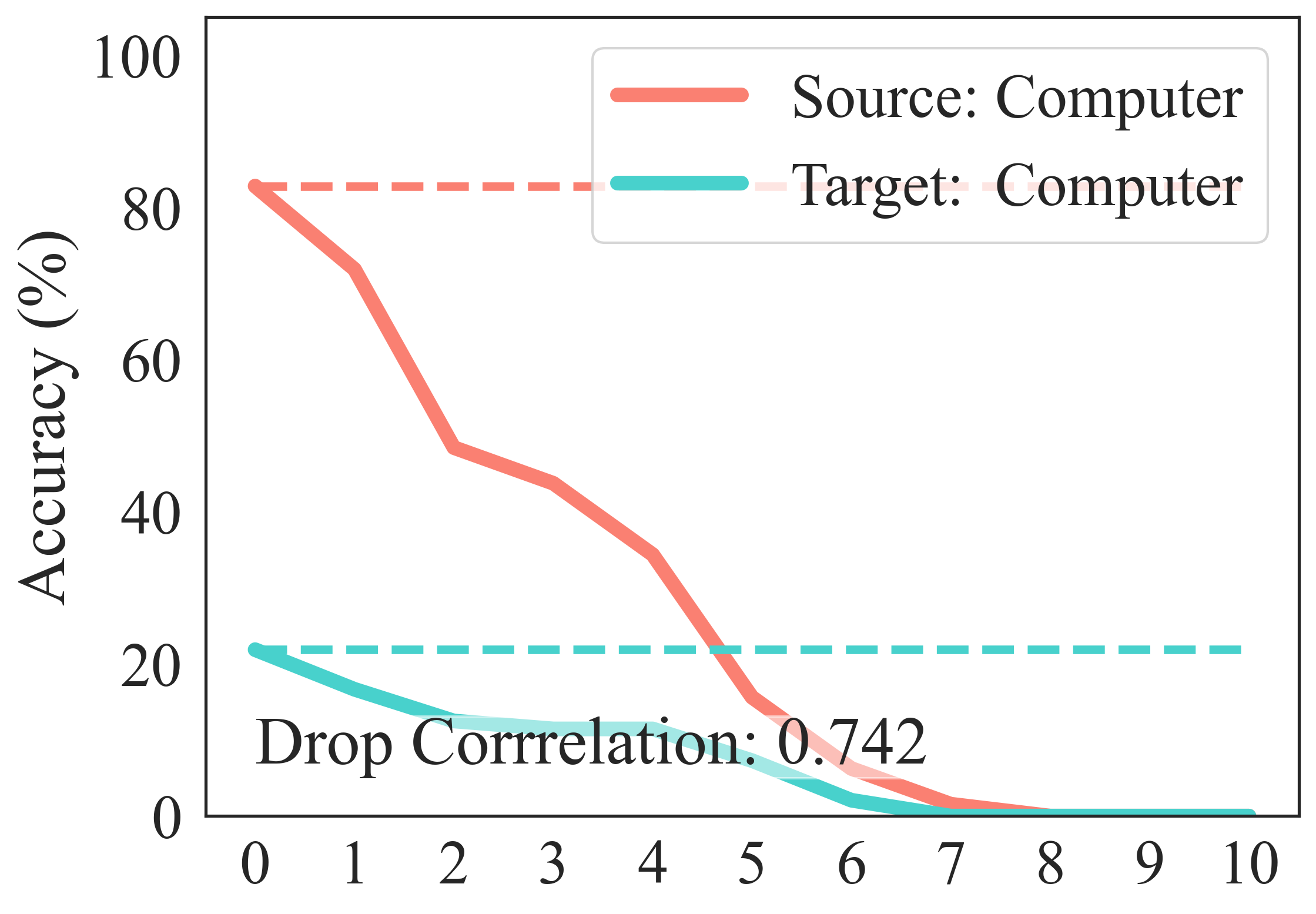}
         \caption{ \hspace{0.1cm} Rw→Pr: \hspace{0.1cm} Computer}
         \label{fig:Rw2PrComputer}\vspace{-1mm}
     \end{subfigure}
     \caption{Predicting accuracy after removing the most related prototypes for each category. (a-c) The overall accuracy of all classes, per-class accuracy of {Flowers} and {Sink} classes in task Pr→Cl on \textit{Office-Home}. (d-e) The overall accuracy of all classes, per-class accuracy of {Backpack} and {Computer} classes in task Rw→Pr on \textit{Office-Home}.}
    \label{fig:removed}\vspace{-3.5mm}
\end{figure*}

\noindent \textbf{Implementation}. We implement\footnote{Our code is available at \href{https://github.com/visualizing-transferred-knowledge/vsualizing-transferred-knowledge}{\textit{https://github.com/visualizing-transferred-knowledge/vsualizing-transferred-knowledge}}.} our model using PyTorch\cite{NEURIPS2019_9015} with one NVIDIA TITAN RTX GPU. We use the ResNet-34~\cite{he2016deep} pre-trained on \textit{ImageNet} as the backbone feature extractor and train the base domain adaptation (DA) models on all the datasets. All images are resized to 224$\times$224. We use the PyTorch built-in random horizontal flip as the on-the-fly data augmentation. After the base DA model is trained, we fix the backbone feature extractor and the feature classifier. Then, we train our prototype layers to extract prototypes from the domain-invariant feature space and align the prototype classifier with the feature classifier. The dimension of the output of the ResNet-34 backbone is $H$$=$$W$$=$$7$ with 128 channels, and the size of each prototype is set to $1$$\times$$1$$\times$$128$ accordingly. The number of prototypes for each category is set to 10 for all datasets. The learning rate is set to 0.003 for both the prototype layer and the prototype classifier. $\alpha$, $\beta$, and $\lambda$ are set to 0.8, 10, and 0.0001 in all the experiments following the same settings as ProtoPNet. For \textit{Office-Home}~\cite{venkateswara2017deep} dataset, $\gamma$ is set to 100 and for \textit{DomainNet-126}~\cite{peng2019moment}, we set $\gamma$ to 10. On both datasets, we train the model for 100 epochs and push the prototypes every 10 epochs followed by last layer optimization. After each push stage, we visualize the transferred prototypes by identifying their closest source image parts.

\subsection{Algorithmic performance}
We conduct experiments of our interpretative model for DANN~\cite{ganin2016domain} with ResNet-34~\cite{he2016deep} on the benchmarks including \textit{Office-Home} and \textit{DomainNet-126}. As we discussed in \Cref{sec:intro}, the motivation of our work is to interpret the CNN-based DA models, and \textit{faithfully} visualize the transferred knowledge rather than pursuing higher performance.

As shown in \Cref{table:dann_OfficeHome}, the performance of our interpretive model is consistent with DANN on the target domain in terms of predicting accuracy. Averagely, our model only deteriorates by 0.17\% and the accuracy differences are within 1\% for all 12 tasks on \textit{Office-Home} dataset. This tiny performance gap indicates that our model keeps loyal to the DANN model when making predictions, and the prototypes learned from the share feature space capture most of the information used by the base model. This high fidelity enables our model to provide a case-based interpretation for the DANN model. 
We also conduct experiments on \textit{DomainNet-126} dataset and report the results in the same table. On this large-scale dataset, our model diverges further from the base DA model on all 12 tasks, averaging a 4.24\% accuracy drop. We conjure that aligning the prototype classifier $f_p$ will be considerably more difficult on this complex dataset with 126 categories, hindering our model's ability to emulate the base model.

\begin{figure*}[t]
     \begin{subfigure}[b]{0.66\textwidth}
         \centering
         \includegraphics[width=\textwidth]{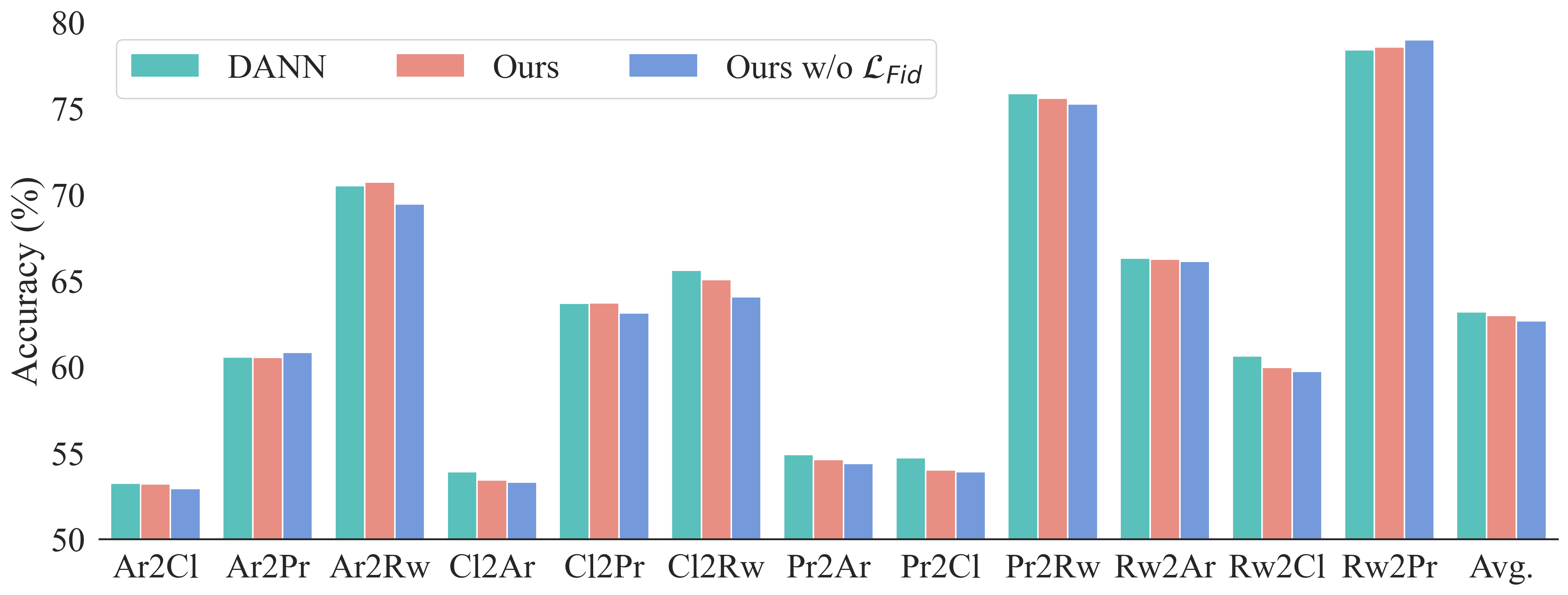}
         \caption{Ablation study of the knowledge fidelity preservation loss $\mathcal{L}_{Fid}$ with DANN model}
         \label{fig:fid}\vspace{-2mm}
     \end{subfigure}
     \hfill
     \begin{subfigure}[b]{0.313\textwidth}
         \centering
         \includegraphics[width=\textwidth]{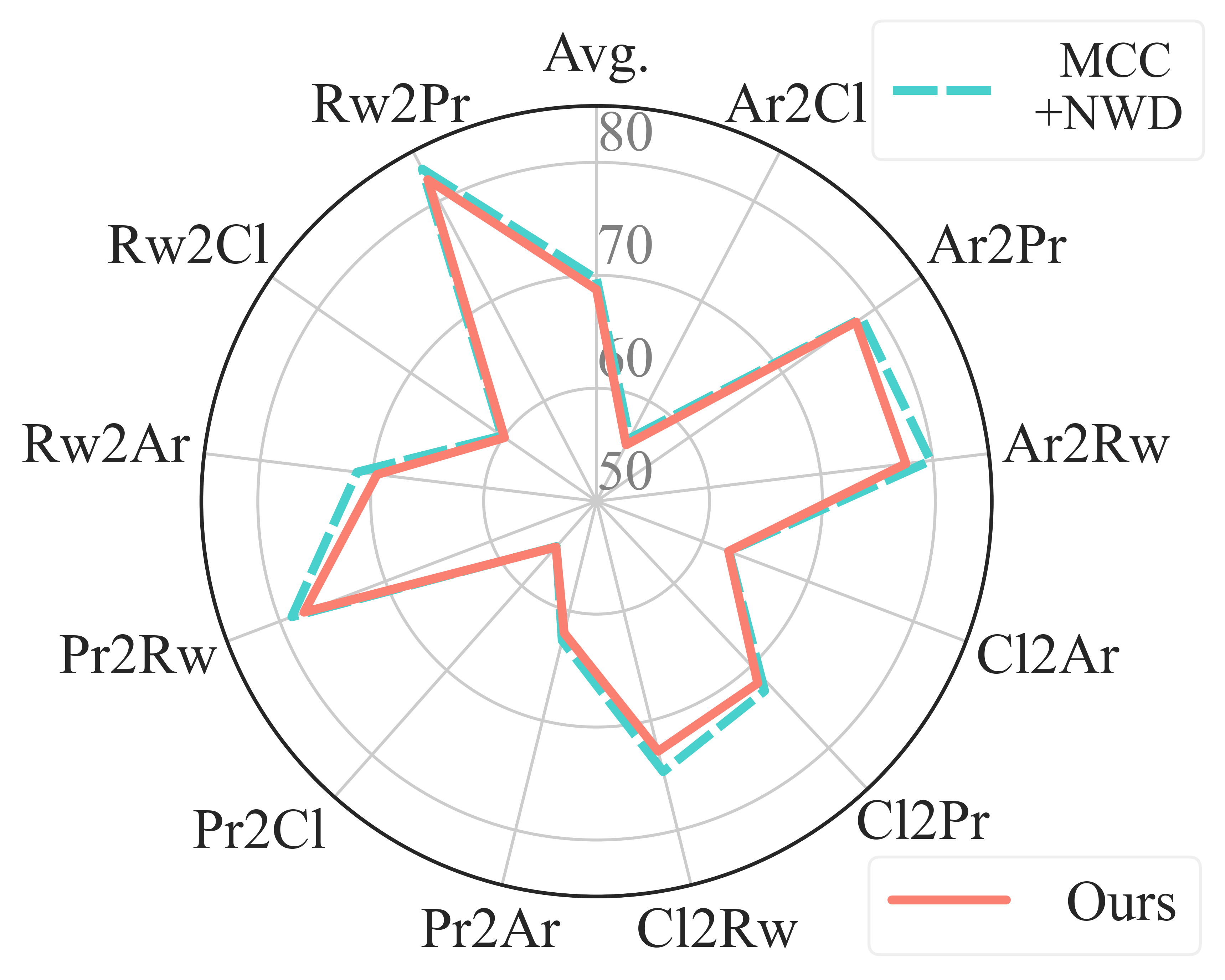}
         \caption{Our interpretation on MCC+NWD}
         \label{fig:backbone}\vspace{-2mm}
     \end{subfigure}

     \caption{(a) Bar plot for comparing our full model with our model without $\mathcal{L}_{Fid}$ when interpreting DANN~\cite{ganin2016domain} base model on \textit{Office-Home} dataset. (b) Radar plot of our interpretation on MCC+NWD~\cite{9879990} model on \textit{Office-Home} dataset.}
    \label{fig:ablation}\vspace{-3mm}
\end{figure*}

\subsection{In-depth exploration}

\noindent \textbf{Prototypical visual matching.} To check the transferred knowledge for each category during the adaptation, we select two categories from the task Pr→Cl of \textit{Office-Home} dataset and visualize prototypes learned from the domain-invariant feature space. The overall accuracy of our model for this task is 54.02\% across all categories, in which we choose {Flowers} and {Sink} as examples. The source and target predicting accuracy of {Flowers} are 98.90\% and 88.89\% respectively, indicating the model successfully transferred the knowledge learned from the source domain to the target domain. However, the source accuracy of {Sink} is 100.00\% while the accuracy on the target domain is merely 2.38\%, implying the model does not adapt well for this category. 

In order to understand this vast performance difference, we look into the most related prototypes for {Flowers} and {Sink} categories on both source and target images as examples. In \Cref{fig:matching}, we show the {Flowers} prototypes on the left and the {Sink} prototypes on the right. For each prototype, we crop out each prototype on the original source sample, then find three example patches associated with it in both domains. For {Flowers} class, the model extracts prototypes containing various information from the source images and successfully transfers the knowledge to the target domain. As shown in \Cref{fig:matching} (a-c), we can intuitively interpret these three prototypes as green stems and petals with different biological forms. These three prototypes all match the corresponding information across the source and target samples, indicating that diverse transferred knowledge is involved in predicting the target samples. These well-transferred prototypes could bolster users' confidence for the {Flowers} category, as some of the most distinguished traits are recognized across domains.

On the other hand, the {Sink} prototypes are less informative for transferring knowledge to the target domain. The first two prototypes in \Cref{fig:matching}~(d-e), presumably representing the edge and hose of a sink according to the source patches, connect some target samples with similar semantics but also point to background areas, which we crop out with red rectangles. On top of that, the last prototype in \Cref{fig:matching}~(f) almost only focuses on noisy parts of the target samples, suggesting that only limited and less informative knowledge is being transferred. The observations on the mismatched information are just as important because they ring up the curtain for diagnosing and improving the base model. 

\noindent \textbf{Prototype inspections.} In addition to visualization, we conduct experiments to further examine the transferability of the discovered prototypes on two tasks from the \textit{Office-Home} dataset. We remove the most category-related prototypes one by one, by masking their corresponding weight in the prototypical classifier $h_p$, and plot the predicting accuracy on both domains after each removal step in \Cref{fig:removed}. 
Moreover, we calculated Spearman's correlation coefficient between the performance drop on the source and target domain after each removing step, to check if the prototypes indeed contain the transferred knowledge. Ideally, the accuracy on two domains should be in a similar trend and highly correlated if each prototype contains knowledge that facilitates the prediction of both domains. As shown in \Cref{fig:removed} (a) and \Cref{fig:removed} (d), the overall accuracy for all classes synchronously decreases in both domains, indicating our prototypes carry the matched knowledge transferred during adaptation. Their correlations of the performance drop also validate the knowledge transfer. For individual classes, we choose two classes for each task, where our model performs quite differently on the target domain. As expected, the {Flowers} and {Backpack} classes show very similar trends in terms of performance drop after removing prototypes, which accords with their high target accuracy. This implies sufficient knowledge gets transferred to the unseen domain. The {Sink} and {Computer}, on the contrary, only achieve 2.38\% and 21.88\% on the target domain. The low performance can be explained by their contrasting accuracy plots and low correlations, which underlines the fact that insufficient meaningful information is transferred for predicting target samples. Notice this observation for {Flowers} and {Sink} categories on task {Pr→Cl} concurs with our visualization in the last section, i.e., the knowledge transferred to the target {Sink} class carries a great amount of misinformation. Combining the prototypical visual matching and prototype inspections, our method provides a comprehensive interpretation of the \textit{transferred knowledge} in the DA models, setting a solid cornerstone for the downstream tasks.

\noindent \textbf{Ablation study.} We perform two additional experiments in our ablation study. Firstly, we remove the knowledge fidelity preservation from our model and compare its predicting accuracy against DANN~\cite{ganin2016domain} and our full model on \textit{Office-Home}. As shown in \Cref{fig:ablation} (a), adding $\mathcal{L}_{Fid}$ slightly improves the model in terms of being loyal to the base model. With this fidelity loss term, our model performs closer to the original DANN in all tasks, resulting in a 3.1\% average improvement. This enhancement demonstrates that knowledge fidelity preservation indeed helps our prototype catches the transferred knowledge and helps the interpretative classifier imitate the base model's behavior, despite that our training does not directly involve the ground truth labels. Besides, we also apply our interpretative model for another base DA model, NWD~\cite{9879990}, which achieves state-of-art on \textit{Office-Home} combined with MCC~\cite{jin2020minimum}. We use ResNet-34~\cite{he2016deep} as the backbone feature extractor and conduct experiments of MCC+NWD on \textit{Office-Home} with the same settings as the original paper. The results of the original MCC+NWD and our interpretative model are plotted in \Cref{fig:ablation} (b). Our model can also achieve comparable accuracy in all tasks on \textit{Office-Home} dataset with the MCC+NWD model, averaging a slight performance drop of 0.23\%. This small margin indicates the effectiveness of our model in extracting transferred knowledge for various base DA models.

\section{Conclusion}
In this paper, we proposed an interpretative method of unsupervised domain adaptation to visualize the transferred knowledge. To the best of our knowledge, this paper is the first work trying to untangle the domain process with visual illustrations. Our method learns prototypes associated with source image patches for the domain-invariant feature space and guides the prototypes to focus on the transferred knowledge with prediction calibration and knowledge fidelity preservation modules. These prototypes can then establish the association between the source and target samples with visual examples. This level of interpretation not only assures the users to rely on the DA models when the transfer is favorable but also helps diagnose the model by pinpointing the negative transfer. The efficacy of our method was manifested by experiments on two benchmark datasets, and we also conducted extensive in-depth explorations to validate our interpretations of the base DA models. 

\bibliographystyle{unsrt}  
\bibliography{references}

\end{document}